\documentclass{article}
\usepackage[preprint, nonatbib]{neurips_2023}
\usepackage{url}
\usepackage{array}
\usepackage{diagbox}
\usepackage{booktabs}
\usepackage{caption}
\usepackage{amssymb}

\title{Proximal Policy Optimization with Adaptive Exploration}
\author{Andrei Lixandru}

\begin{document}

\maketitle

\begin{abstract}
    Proximal Policy Optimization with Adaptive Exploration (axPPO) is introduced as a novel learning algorithm. 
    This paper investigates the exploration-exploitation tradeoff within the context of reinforcement learning and aims to contribute new insights into reinforcement learning algorithm design. 
    The proposed adaptive exploration framework dynamically adjusts the exploration magnitude during training based on the recent performance of the agent. Our proposed method outperforms standard PPO algorithms in learning efficiency, particularly when significant exploratory behavior is needed at the beginning of the learning process.
\end{abstract}

\section{Introduction}

The dual control problem, central to reinforcement learning (RL), poses a fundamental challenge in achieving optimal decision-making. It revolves around striking a balance between exploration, for uncovering new information, and exploitation, for maximizing accumulated rewards based on existing knowledge. In the realm of RL, where agents navigate dynamic environments through sequential decisions, addressing this delicate trade off is crucial for optimizing strategy. The investigation of the dual control problem holds significance for AI research, contributing to the refinement of RL algorithms and the development of intelligent systems capable of robust decision-making in intricate and uncertain scenarios. This paper is part of the larger effort to unveil strategies that can elevate AI systems beyond current limitations, enhancing their adaptability to real-world complexities and uncertainties.

Previous methods to address the dual control problem include curiosity-driven reinforcement learning \cite{still2012information} and the use of the epsilon coefficient in Deep Q Networks \cite{mnih2015human}. Additionally, Proximal Policy Optimization (PPO) algorithms \cite{schulman2017proximal} incorporate an entropy coefficient to incentivize exploration. However, the static nature of this coefficient throughout training poses limitations. This paper proposes an adaptive exploration framework within the context of PPO, aiming to dynamically scale the entropy coefficient based on the agent's recent performance.

In short, our adaptive exploration framework works by scaling the entropy coefficient of the classic PPO algorithm by the recent return received by the agent, previous to the current time step in the learning process. A more in-depth description of the adaptive exploration framework follows in the Methods section.
\section{Methods}

\subsection{Algorithm}
The methods for adaptive exploration we developed are used in the context of PPO algorithms. A traditional objective of a PPO algorithm is the one introduced by OpenAI \cite{schulman2017proximal}, represented by the following loss function:

\begin{equation}
    L_t(\theta) = \hat{\mathbb{E}}_t\left[L_t^{CLIP}(\theta) - c_1 L_t^{VF}(\theta) + c_2 S[\pi_\theta](s_t)\right]
\end{equation}

Our focus is on the entropy bonus represented by $S$ and the entropy coefficient $c_2$, which determines the exploration magnitude. In traditional PPO implementations, $c_2$ remains a fixed hyperparameter throughout training. This paper introduces a new learning algorithm, PPO with adaptive exploration (axPPO), which incorporates a dynamic scaling of the entropy coefficient based on the recent return ($G_{recent}$) obtained by the agent:

\begin{equation}
    L_t(\theta) = \hat{\mathbb{E}}_t\left[L_t^{CLIP}(\theta) - c_1 L_t^{VF}(\theta) + G_{recent} \,c2 \,S[\pi_\theta](s_t)\right]
\end{equation}

The adaptive exploration framework relies on $G_{recent}$, computed as:

\begin{equation}
    G_{recent} = G_t(\tau) = \frac{1}{G_{max}}\frac{\sum_{i=t-\tau}^{t} \overline{G}_i^{batch}}{\tau}
\end{equation}

Here, $t$ represents the current time step, $\tau$ is a time constant determining how far back to integrate past returns, and $\overline{G}_i^{batch}$ is the mean return across batches at time step $i$. The result is scaled between 0 and 1 by dividing it by $G_{max}$, the maximum return an agent can receive during an episode. In short, $G_{recent}$ represents the recent performance of the agent, parameterized by $\tau$.

\subsection{Experiments}
In our experiments, we will draw a comparison between the performance of a standard PPO algorithm, with the loss function from eq. (1) and the performance of PPO with adaptive exploration, with the loss function from eq. (2). We will compare their return across different entropy coefficients that range from 0 to 0.8. We will also include a comparison of axPPO algorithms across a range of time constants, ranging from 1 to 200.

In short, the research question we are answering is: Does a PPO algorithm benefit from adaptive exploration that is based on recent agent performance?

For each experimental condition, we will perform 3 runs and include only the mean over the runs in the Performance table from the Results section. All agents are trained in the Gymnasium CartPole-v1 environment \cite{towers_gymnasium_2023}, for 60,000 learning steps. An actor-critic setup is used for each condition, with a multilayer perceptron of 2 hidden layers with 64 units per layer, the parameters of the actor and critic being shared. The software for running the experiments is freely available in the form of a Python notebook \cite{repo}.

\section{Results}

\begin{table}[h]
  \centering
  \caption{Performance table} 
  \vspace{\baselineskip}

  \label{tab:example}
  \begin{tabular}{*{6}{c}}
    \toprule
    \diagbox[width=17.5em]{Algorithm}{Entropy Coefficient} & 0 & 0.1 & 0.3 & 0.5 & 0.8 \\
    \midrule
    Standard PPO & 455 & 430 & 340 & 211 & 111 \\
    axPPO $\tau$ = 1 & - & 459 & 443 & 447 & 423 \\
    axPPO $\tau$ = 10 & - & 453 & 459 & 449 & 406 \\
    axPPO $\tau$ = 20 & - & 453 & 455 & 439 & 413 \\
    axPPO $\tau$ = 50 & - & 459 & 450 & 444 & 433 \\
    \textbf{axPPO $\tau$ = 100} & - & \textbf{462} & 446 & 429 & 425 \\
    axPPO $\tau$ = 200 & - & 458 & 449 & 440 & 409 \\
    \bottomrule
  \end{tabular}
\end{table}

Table 1 contains the return collected during an episode after 60,000 steps of training. Columns represent different values of the entropy coefficient. Values for a coefficient of 0 for axPPO are omitted, as for values of 0, axPPO functions exactly as standard PPO.

The performance table indicates that axPPO has a competitive performance when compared to standard PPO, even surpassing it for some values of $\tau$. 

A surprising finding from the performance table is that axPPO does not lose significant learning speed when increasing the entropy coefficient. This contrasts with the performance of standard PPO, for which the reward collected with entropy coefficient is 0.8 is a forth of the reward collected when the entropy coefficient is 0.1.

When comparing the performance of asPPO over different values of $\tau$, it can be seen that the effectiveness of exploration adapted to the current reward ($\tau=1$) is comparable to exploration based on the recent reward ($\tau \in \{10, 20, 50, 100, 200\}$). Integrating a memory of past returns in the loss function does not seem to provide additional value compared to using the current return.

\section{Discussion}

The performance of axPPO demonstrates promise, although its definitive superiority over standard PPO remains uncertain. However, it is evident that axPPO exhibits a distinct advantage over standard PPO in scenarios where substantial exploratory behavior is essential or beneficial during the initial phases of the learning process (i.e., high entropy coefficients). This implies that having an agent with a dynamic exploratory behavior that is dependent on its performance fosters more efficient exploration. Thus, axPPO opens a new direction for tackling the exploration-exploitation trade off from reinforcement learning.

A more comprehensive comparative analysis between standard PPO and axPPO is imperative, encompassing assessments in more intricate environments with expansive action spaces, such as Atari games or robotics environments. 

The incorporation of adaptive exploration has proven to enhance the learning trajectory of Proximal Policy Optimization algorithms, ensuring sustained performance across varying magnitudes of initial exploration. Further investigations and rigorous evaluations are warranted to establish the nuanced dynamics and potential advantages of axPPO in diverse and complex settings.

\section{Conclusion}

In conclusion, the adaptive exploration framework presented in axPPO demonstrates to be superior to standard PPO in maintaining performance across different magnitudes of initial exploration. The findings highlight the potential advantages of dynamically adjusting the exploration term based on the agent performance in reinforcement learning algorithms. Further research and comprehensive comparisons are recommended to fully understand the capabilities and limitations of axPPO.

% Your references here
%\bibliography{references.bib}

\bibliographystyle{unsrt}

\end{document}